\documentclass[10pt,twocolumn,letterpaper]{article}

\usepackage{cvpr}              

\usepackage{graphicx}
\usepackage{amsmath}
\usepackage{amssymb}
\usepackage{booktabs}
\usepackage{stfloats}
\usepackage{color}
\usepackage{times}
\usepackage[pagebackref,breaklinks,colorlinks]{hyperref}
\usepackage{overpic}
\usepackage{bm}
\usepackage{tabu}
\usepackage{bbding}
\usepackage{multicol}
\usepackage{multirow}
\usepackage{scalerel,stackengine}
\usepackage[table]{xcolor}
\usepackage[accsupp]{axessibility}
\newcommand{\PAR}[1]{\vskip3pt \noindent{\bf #1~}}
\usepackage{array}
\usepackage[capitalize]{cleveref}
\crefname{section}{Sec.}{Secs.}
\Crefname{section}{Section}{Sections}
\Crefname{table}{Table}{Tables}
\crefname{table}{Tab.}{Tabs.}
\usepackage{algorithm}
\usepackage{algorithmic}
\usepackage[pagebackref,breaklinks,colorlinks]{hyperref}

\usepackage[capitalize]{cleveref}
\crefname{section}{Sec.}{Secs.}
\Crefname{section}{Section}{Sections}
\Crefname{table}{Table}{Tables}
\crefname{table}{Tab.}{Tabs.}

\def\cvprPaperID{5726} 
\def\confName{CVPR}
\def\confYear{2023}

\title{
CIMI4D: A Large Multimodal Climbing Motion Dataset\\ under Human-scene Interactions
\vspace{-4mm}
}
\author{Ming Yan$^{1,2,3\ast}$\hspace{4mm} Xin Wang$^{1,3\ast}$\hspace{4mm} Yudi Dai$^{1,3}$\hspace{4mm} Siqi Shen$^{1,3\dagger}$\hspace{4mm} Chenglu Wen$^{1,3}$\\ Lan Xu$^4$\hspace{4mm} Yuexin Ma$^4$\hspace{4mm} Cheng Wang$^{1,3}$\\
$^1$Fujian Key Laboratory of Sensing and Computing for Smart Cities, Xiamen University\\
$^2$National Institute for Data Science in Health and Medicine, Xiamen University\\
$^3$Key Laboratory of Multimedia Trusted Perception and Efficient Computing,\\ Ministry of Education of China, School of Informatics, Xiamen University\\
$^4$Shanghai Engineering Research Center of Intelligent Vision and Imaging, ShanghaiTech University\\
{}
\vspace{-22mm}
\and
\\
{}
}

\begin{document}
\twocolumn[{%
\renewcommand\twocolumn[1][!htb]{#1}%
\maketitle
\vspace{-6mm}
\begin{center}
    \centering
    \includegraphics[width=1\linewidth]{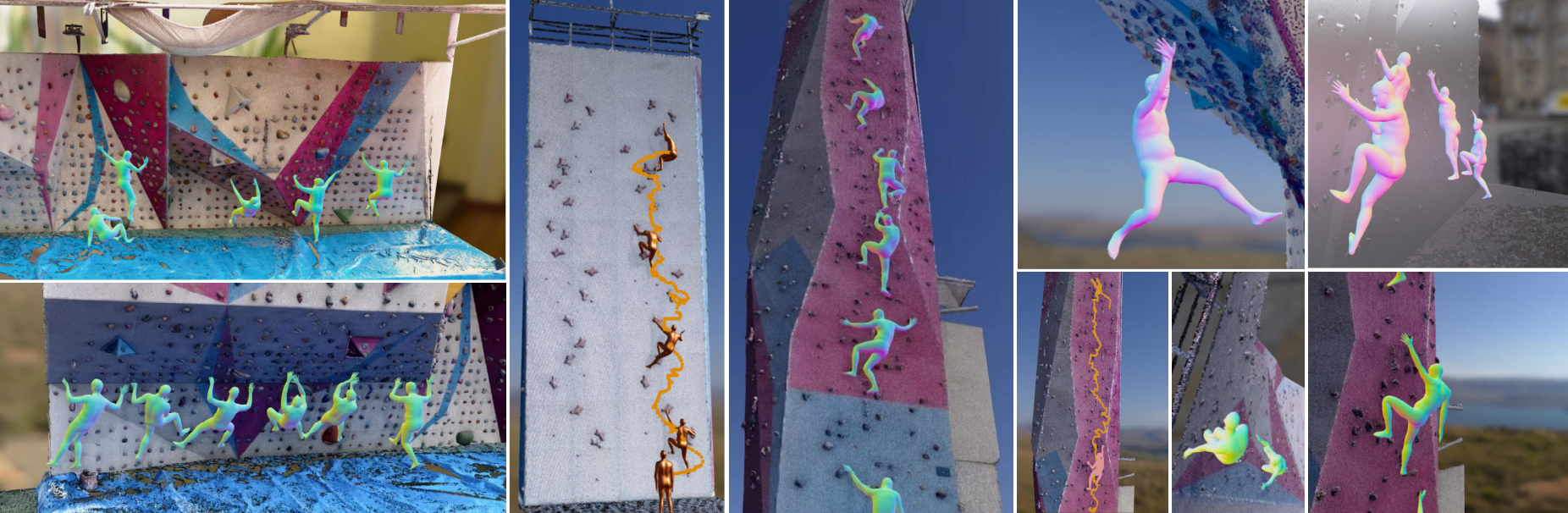}
    \vspace{-6mm}
    \captionof{figure}{CIMI4D is a dataset of rock climbing motions recorded using RGB cameras, LiDAR, and IMUs. CIMI4D contains 42 action sequences of 12 actors climbing 13 climbing walls, and provides finely annotated human poses and global trajectories (orange lines).  The pictures showcase various types of complex scenes (7 of them has high-precision point cloud scan) and challenging actions in CIMI4D. }
    \label{fig:gallery_climbing}
\end{center}%
}]


\footnotetext{$^\ast$ Equal contribution.}
\footnotetext{$^\dagger$ Corresponding author.}
\begin{abstract}
\vspace{-4mm}
Motion capture is a long-standing research problem. Although it has been studied for decades, the majority of research focus on ground-based movements such as walking, sitting, dancing, etc. Off-grounded actions such as climbing are largely overlooked. As an important type of action in sports  and firefighting field, the climbing movements is challenging to capture because of its complex back poses, intricate human-scene interactions, and difficult global localization. The research community does not have an in-depth understanding of the climbing action due to the lack of specific datasets. To address this limitation, we collect CIMI4D, a large rock \textbf{C}l\textbf{I}mbing \textbf{M}ot\textbf{I}on dataset from 12 persons climbing 13 different climbing walls. The dataset consists of around 180,000 frames of pose inertial measurements, LiDAR point clouds, RGB videos, high-precision static point cloud scenes, and reconstructed scene meshes. Moreover, we frame-wise annotate touch rock holds to facilitate a detailed exploration of human-scene interaction. The core of this dataset is a blending optimization process, which corrects for the pose as it drifts and is affected by the magnetic conditions. To evaluate the merit of CIMI4D, we perform four tasks which include human pose estimations (with/without scene constraints), pose prediction, and pose generation. The experimental results demonstrate that CIMI4D presents great challenges to existing methods and enables extensive research opportunities. We share the dataset with the research community in \href{http://www.lidarhumanmotion.net/cimi4d/}{http://www.lidarhumanmotion.net/cimi4d/}. 


\end{abstract}

\vspace{-5mm}
\section{Introduction}
\label{sec:intro}
Capturing human motions can benefit many downstream applications, such as AR/VR, games, movies, robotics, etc. However, it is a challenging and long-standing problem~\cite{Martinez17,alldieck2017optical,SMPL2015,Zhou16a,OpenPose} due to the diversity of human poses and complex interactive environment. Researchers have proposed various approaches to estimate human poses from images~\cite{SPIN_ICCV2019,MonoPerfCap,DeepCap_CVPR2020,challencap,LiveCap2019tog}, point clouds~\cite{lidarcap}, inertial measurement units (IMUs)~\cite{DIP,PIP}, etc. Although the problem of human pose estimation (HPE) has been studied for decades~\cite{Bregler1998TrackingPW,Vlasic2007PracticalMC,woltring1974new}, most of the existing solutions focus on upright frontal poses on the ground (such as walking, sitting, jumping, dancing and yoga)~\cite{survey2022}. Different from daily activities (such as walking and running) that are on the ground, climbing is an activity off the ground with back poses, which is also an important type for sports~\cite{ViconForceClimbing,exemPose}, entertainment~\cite{MRClimbing17,ClimbingMR,footVR}, and firefighting. 

Climbing is an activity that involves ascending geographical objects using hands and feet, such as hills, rocks, or walls. Estimating the pose of a climbing human is challenging due to severe self-occlusion and the human body's closely contact with the climbing surface. These issues are primarily caused by complex human-scene interactions. Moreover, understanding the climbing activities requires both accurate captures of the complex climbing poses and precise localization of the climber within scenes, which is especially challenging. Many pose/mesh estimation methods are data-driven methods~\cite{HMR,HUMOR_ICCV2021,survey2022,ICON}, relying on huge climbing motion data for training networks. So a large-scale climbing dataset is necessary for the holistic understanding of human poses. Publicly available human motion datasets are mostly in upright frontal poses\cite{climbingSurvey2020,ClimbingSensorSurvey2022,AMASS_ICCV2019}, which are significantly different from climbing poses. Albeit some researchers collected RGBD-based climbing videos~\cite{Beltrn2022AutomatedHM} or used marker-based systems~\cite{ViconForceClimbing}, their data is private and the scale of dataset is very limited.



To address the limitations of current datasets and boost related research, we collect a large-scale multimodal climbing dataset, CIMI4D, under complex human-scene interaction, as depicted in~\cref{fig:gallery_climbing}. CIMI4D consists of around 180,000 frames of time-synchronized and well-annotated LiDAR point clouds, RGB videos, and IMU measurements from 12 actors climbing 13 rock-climbing walls. 12 actors include professional athletes, rock climbing enthusiasts, and beginners. In total, we collect 42 rock climbing motion sequences, which enable CIMI4D to cover a wide diversity of climbing behaviors. To facilitate deep understanding for human-scene interactions, we also provide high-quality static point clouds using a high-precision device for seven rock-climbing walls. Furthermore, we annotate the rock holds (holds) on climbing walls and manually label the contact information between the human body and the holds. To obtain accurate pose and global trajectory of the human body, we devise an optimization method to annotate IMU data, as it drifts over time~\cite{3DPW,dou-siggraph2016} and is subject to magnetic conditions in the environment.


The comprehensive annotations in CIMI4D provide the opportunity for benchmarking various 3D HPE tasks. In this work, we focus on four tasks: human pose estimation with or without scene constraints, human pose prediction and generation. To assess the effectiveness of existing methods on these tasks, we perform both quantitative and qualitative experiments. However, most of the existing approaches are unable to capture accurately the climbing action. Our experimental results demonstrate that CIMI4D presents new challenges for current computer vision algorithms.
We hope that CIMI4D could provide more opportunities for a deep understanding of human-scene interactions and further benefit the digital reconstruction for both. In summary, our contributions can be listed as below:

\vspace{-2mm}
\begin{itemize}
    \item We present the first 3D climbing motion dataset, CIMI4D, for understanding the interaction between complex human actions with scenes. CIMI4D consists of RGB videos, LiDAR point clouds, IMU measurements, and high-precision reconstructed scenes.
    \vspace{-1mm}
    \item  We design an annotation method which uses multiple constraints to obtain natural and smooth human poses and trajectories. 
    \vspace{-1mm}
    \item We perform an in-depth analysis of multiple methods for four tasks. CIMI4D presents a significant challenge to existing methods.
\end{itemize} 
\vspace{-2mm}

\label{sec:intro}

\section{Related Work}\label{sec:related}

\subsection{Human Pose Datasets}


The focus of human pose estimation research is partially driven by the design of datasets. To recover 2D poses from RGB videos, researchers have proposed various datasets~\cite{Zhang2013FromAT,Andriluka2018PoseTrackAB,Carreira2017QuoVA,Kanazawa2019Learning3H}. For 3D human pose estimation, researchers have collected multiple datasets \cite{Sigal2009HumanEvaSV,Ionescu2014Human36MLS,Trumble2017TotalC3,AMASS_ICCV2019,AGORA}.

HumanEva\cite{Sigal2009HumanEvaSV} contains 4 subjects performing a set of predefined actions within indoor scenarios, and with static background. 
The Human3.6M\cite{Ionescu2014Human36MLS} consists of human poses from 11 actors within 17 controlled indoor scenarios. The scenarios consist discussion, greeting, walking, waiting, eating, sitting, etc. Its 3D ground truth is collected through marker-based approaches. 
MPI-INF-3DHP\cite{Mehta2017Monocular3H} captures human motion using a multi-camera markerless motion capture system in a green screen studio. It consists 8 actors performing 8 activities including walking/standing, sitting/reclining, exercising/crouching, dancing/sports. Except the diving activities, most of the activities are ground-based activities. 
TotalCapture\cite{Trumble2017TotalC3} provides a 3D human pose dataset consists of synchronized multi-view videos and IMU. It is collected in a green scene studio wherein actors perform actions such as walking, running, yoga, bending, crawling, etc.  
3DPW\cite{3DPW} is an in-the-wild 3D dataset collected through a set of IMU sensors and a hand-held camera. It contains 51,000 video frames of several outdoor and indoor activities performed by 7 actors. 
PedX\cite{Kim2019PedXBD} consists of 5,000 pairs of stereo images and LiDAR point clouds for pedestrian poses. AMASS\cite{AMASS_ICCV2019} is a large-scale MoCap dataset, which spans over 300 subjects and contains 40 hours of motion sequences. The LiDARHuman26M\cite{lidarcap} consists of LiDAR point clouds, RGB videos, and IMU data. It records 13 actors performing 20 daily activities in 2 controlled scenes.
To our best knowledge, SPEED21~\cite{elias2021speed21} is the only published climbing dataset. It labels climbing athletes' 2D joints from sport-events videos (speed climbing only). CIMI4D (120 minutes) is larger than SPEED21 (38 minutes) and has multiple modalities with 3D scenes. 

\subsection{Human Pose Datasets with Scene Constraints}
PROX~\cite{PROX} records human-scene interactions in a variety of indoor scenes through a RGBD camera. Each indoor scenes are pre-scanned using Structured RGBD scanners. 4DCapture\cite{MiaoLiu20204DHB} collects egocentric videos to reconstruct second-person 3D human body meshes without reliable 3D annotations. HPS\cite{HPS} uses IMUs and head-mounted cameras to reconstruct human poses in large 3D scenes, but does not interact with the scene. EgoBody\cite{EgoBody} records human-interaction from egocentric views. HSC4D\cite{HSC4D} is a human-centered 4D scene capture dataset for human pose estimation and localization. It is collected by body-mounted IMU and LiDAR through walking in 3 scenes. RICH\cite{RICH} contains multiview outdoor/indoor video sequences, ground-truth 3D human bodies, 3D body scans, and high resolution 3D scene scans.


\subsection{Pose Estimation Methods}


Extensive work has focused on estimating the pose, shape, and motion of human from pure vision-base data~\cite{PARE_ICCV2021,MAED}. PiFu\cite{pifu}, PiFuHd~\cite{pifuhd} and ICON~\cite{ICON} estimate clothed human from RGB images. GLAMR~\cite{GLAMR} estimate global human mesh with dynamic cameras. RobustFusion~\cite{Su2020RobustFusionHV}, EventCap~\cite{Xu2020EventCapM3} and LiDARCap~\cite{lidarcap} capture human motion using a RGBD camera, an event camera and a LiDAR, respectively. FuturePose~\cite{futurepose} predicts the movement of skeleton human joints.  S3\cite{yang2021s3} and TailorNet~\cite{Patel2020TailorNetPC} represent human pose, using neural implicit function. 

Human pose priors are used in pose estimation tasks~\cite{LEMO,HMR,VIBE,VPose,HUMOR_ICCV2021,posendf,DynaBOA}. Most of them learn priors from the AMASS dataset~\cite{AMASS_ICCV2019}. Due to the lack of datasets, only a few work considers human scene interactions, PROX~\cite{PROX} and LEMO~\cite{LEMO} estimate human poses with scene constraints. POSER~\cite{POSER} populates scenes with realistic human poses. Besides the vision-based approaches, body-worn IMUs\cite{pons2011outdoor,Yi2021TransPoseR3,SIP,DIP,3DPW,HPS, HSC4D} are used in human pose estimation. Our dataset can be used for developing better human pose estimation methods using different modalities with scenes.

Researchers have use various methods to study the climbing activity~\cite{climbingSurvey2020,ClimbingSensorSurvey2022}. ~\cite{replicating} uses multi-view stereo to reconstruct a rock wall. \cite{pandurevic2022analysis} uses OpenPose~\cite{OpenPose} to extract the skeleton of climbers. \cite{reveret20203d} captures poses and positions through RGB video and a marker.

\label{sec:relatedWork}

\begin{table*}[!t]
    \vspace{-3mm}
     \centering
     \resizebox{\linewidth}{!}{
     \begin{tabular}{lcccccccccc}
        \toprule
        Dataset & 3D scene & Body point & Interaction & LiDAR & RGB video & Trajectory & IMU & Frames & Motions & In/Out-door  \\
        \midrule
        PROX~\cite{PROX} &    \CheckmarkBold   &       &        &         & \CheckmarkBold     & \CheckmarkBold &   &    20k &    Daily   &  Indoor \\
         LiDARHuman26M\cite{lidarcap} &       &   \CheckmarkBold    &      &   \CheckmarkBold    & \CheckmarkBold     &  & \CheckmarkBold &  184k &   Daily  &   Outdoor    \\

        BEHAVE~\cite{bhatnagar2022behave} &       &        & \CheckmarkBold     &       & \CheckmarkBold     &   & \CheckmarkBold  & 15k &   Daily    &    Indoor    \\
         HPS~\cite{HPS} & \CheckmarkBold     &      &       &       & \CheckmarkBold     &  \CheckmarkBold  & \CheckmarkBold & 7k  & Daily  &    Outdoor    \\

        3DPW~\cite{3DPW} &      &      &      &       & \CheckmarkBold     &    - & \CheckmarkBold & 51k &  Daily     &   Outdoor  \\

        HSC4D~\cite{HSC4D} & \CheckmarkBold     & \CheckmarkBold     &        & \CheckmarkBold    &        &    \CheckmarkBold & \CheckmarkBold & 10k & Daily  &    Outdoor    \\

        AMASS~\cite{AMASS_ICCV2019} &      &   \CheckmarkBold   &       &     &    \CheckmarkBold   &  &   &  2420mins &  Many  &    Both    \\
        RICH~\cite{RICH} &   \CheckmarkBold   &   \CheckmarkBold   &    \CheckmarkBold   &     &    \CheckmarkBold   &  &   &  577k &  Many  &    Both    \\
        
        SPEED21~\cite{elias2021speed21} &       &       &        &         & \CheckmarkBold     &  &   &    46k &    Climbing   &  Both \\
        \textcolor[rgb]{0.066,0.466,0.69}{\textbf{CIMI4D(Ours)}}  & \textcolor[rgb]{0.066,0.466,0.69}\CheckmarkBold     & \textcolor[rgb]{0.066,0.466,0.69}\CheckmarkBold     & \textcolor[rgb]{0.066,0.466,0.69}\CheckmarkBold     & \textcolor[rgb]{0.066,0.466,0.69}\CheckmarkBold     & \textcolor[rgb]{0.066,0.466,0.69}\CheckmarkBold     & \textcolor[rgb]{0.066,0.466,0.69}\CheckmarkBold     & 
        \textcolor[rgb]{0.066,0.466,0.69}\CheckmarkBold 
        &\textcolor[rgb]{0.066,0.466,0.69}{180k}
        &\textcolor[rgb]{0.066,0.466,0.69}{\textbf{Climbing}}    &     \textcolor[rgb]{0.066,0.466,0.69}{\textbf{Both}} \\
        \bottomrule
        \end{tabular}%
    }
     
    \vspace{-3mm}
    \caption{Comparison with existing motion datasets.}
    \vspace{-3mm}
    \label{tab:data_compare}
 \end{table*}
 \begin{figure}[!thb]
    \centering
    \includegraphics[width=1\linewidth]{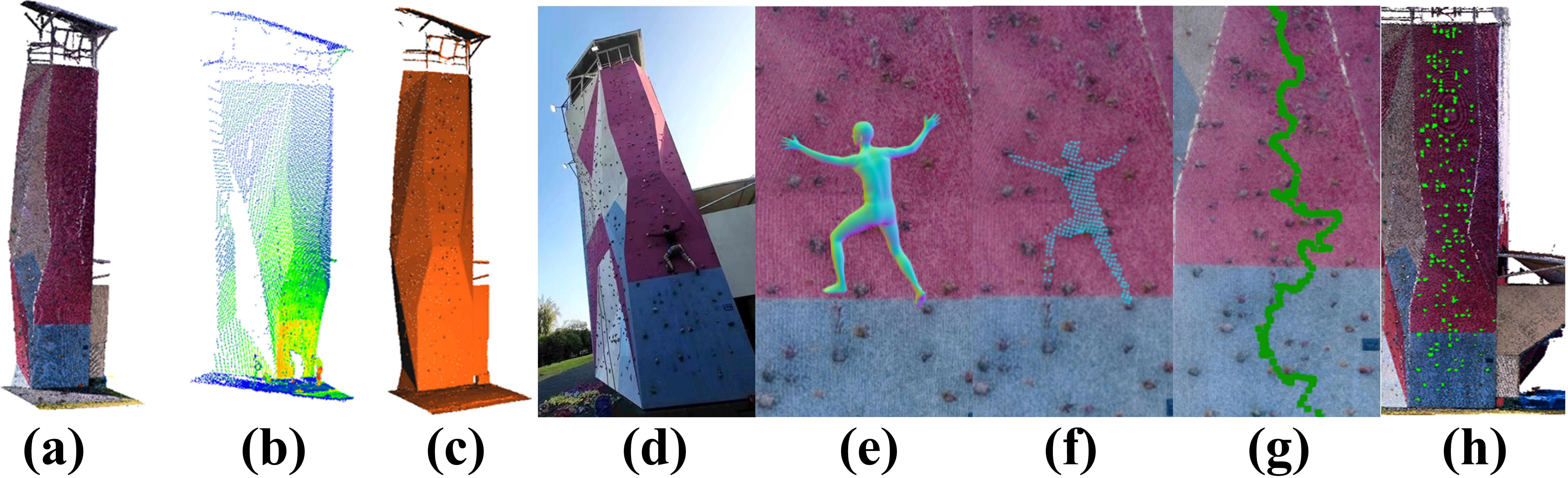}
    \vspace{-5mm}
    \caption{CIMI4D provides rich annotations for different modalities, including (a) High Precision Static Point Cloud, (b) Dynamic Point Cloud Sequence, (c) Reconstructed Mesh Scene, (d) RGB Video, (e) Ground-truth Pose, (f) Body Point Cloud Sequence, (g) Ground-truth Trajectory, (h) Contact (rock holds) Annotation.}
    \vspace{-4mm}
    \label{fig:multimodality}
 \end{figure}

\section{Constructing CIMI4D}\label{sec:dataset}


CIMI4D is a multi-modal climbing dataset that contains 60 minutes of RGB videos, 179,838 frames of LiDAR point clouds, 180 minutes of IMU poses, and accurate global trajectory. \cref{fig:multimodality} depicts different modalities of one scene. Daily activities such as walking and sitting, could be captured in typical rooms with many volunteers available. But capturing the climbing motions should be performed in outdoor or a gym with volunteers having necessary climbing skills. We have invited 12 climbers (professional athletes, enthusiasts, and beginners) to climb on 13 climbing walls. For professional athletes, we do not provide RGB videos. These participants agreed that their recorded data could be used for scientific purposes. In total, we collect 42 complex climbing motion sequences. 

~\cref{tab:data_compare} presents statistics of comparison to other publicly available human pose datasets.  CIMI4D focuses on climbing motions, while most other datasets capture ground movements. Collecting CIMI4D data is more difficult than collecting daily behavior data, as we require setting up multiple devices, including RGB cameras and LiDAR. Each sequence require recalibration of the IMUs to maintain the quality of the dataset. Secondly, our dataset covers multiple modalities, including human points, RGB videos, motion-capture pose data from IMUs, and annotating complex human-scene interactions, which previous datasets did not provide. In addition, CIMI4D includes high-precision 3D LiDAR-scanned point clouds of climbing scenes. Most image-based datasets do not provide depth information or scenes. Finally, CIMI4D consists of the global trajectory of each climber. Most other datasets do not contain global trajectories that are important for human scene understanding.


 \begin{figure}[tb]
    \centering
    \includegraphics[width=1\linewidth]{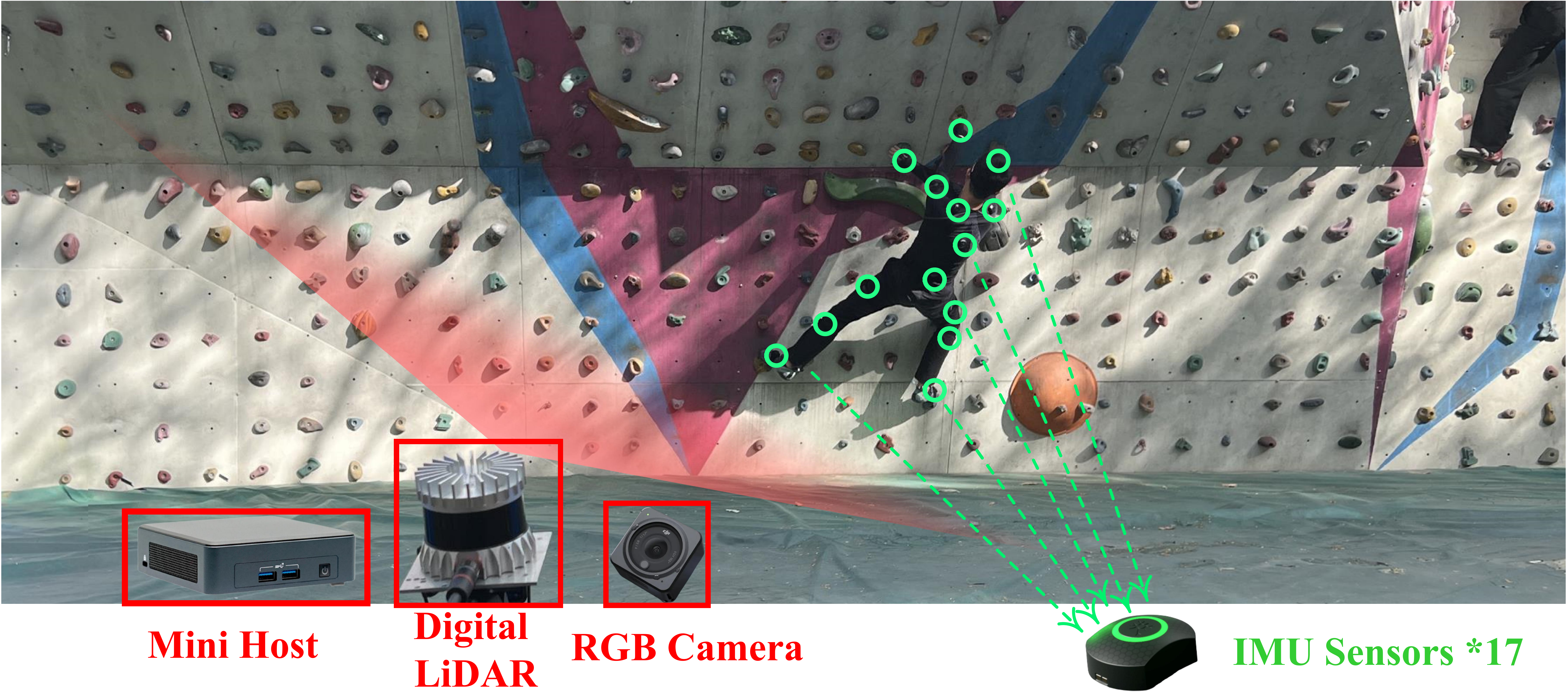}
    \vspace{-6mm}
    \caption{\textbf{Data Capturing System.} Including 17 wearable IMU sensors, a LiDAR, a RGB camera and a computer.}
    \vspace{-4mm}
    \label{fig:Hardware}
 \end{figure}

\subsection{Hardware and Configuration}
\label{sec:Hardware}
To construct CIMI4D, We build a convenient collection system composed of necessary hardware equipment to facilitate our data collection both indoors and outdoors. Every participant wears a Noitom’s inertial MoCap outfit during climbing. As it is depicted in Fig~\ref{fig:Hardware}, each outfit contains 17 IMUs, which records pose data at 100 frame-per-second (FPS). Meanwhile, we use LiDAR (128-beams Ouster-os1) to capture 3D dynamic point clouds at the speed of 20 FPS, and the RGB videos are recorded by an RGB camera (DJI Action 2) at the rate of 60 FPS. The LiDAR has a $360^{\circ}$ horizon and a $45^{\circ}$ vertical field of view, we lay it flat to capture climbers' point clouds for high FOV on vertical walls. The 13 climbing scenes in CIMI4D are categorized into vertical and wide walls for lead climbing, speed climbing(heights up to 20m), and bouldering(long horizontal lengths). We reconstructed seven walls using high-precision RGB 3D point clouds with 40M points, obtained with the Trimble X7 3D laser scanning system.


\PAR{Coordinate Systems.} 
We define three coordinate systems: 1) IMU coordinate system \{$I$\}. 2) LiDAR Coordinate system \{$L$\}. 3) Global/World coordinate system \{$W$\}. We use subscript $k$ to indicate the index of a frame, and superscript, $I$ or $L$ or $W$, to indicate the coordinate system that the data belongs to. For example, the 3D point cloud frames from LiDAR is represented as $P^L = \{P_k^L, k\in~Z^+\}$.

\PAR{Human Pose Model.} A human motion is denoted by $M=(T, \theta, \beta)$, where $T$ represents the $N\times3$ translation parameters, $\theta$ denotes the $N\times24\times3$ pose parameters, and $\beta$ is the $N\times10$ shape parameter following SMPL~\cite{SMPL2015}, $N$ represents the temporal point cloud frames. It use $\varPhi$ to map $(T, \theta, \beta)$ to its triangle mesh model, $V_k, F_k =\varPhi(T, \theta, \beta)$, where body vertices $V_k\in~\mathbb{R}^{6890\times3}$ and faces $F_k\in~\mathbb{R}^{13690\times3}$. 

\PAR{Annotation} The pose $\theta$ and the translation $T$ obtained from the IMU measurements may be inaccurate. IMUs suffer severe drifting for long-period capturing. Further, IMUs subject to the magnetic condition of the environments. We seek to find the precise $T$ and $\theta$ for CIMI4D as annotation labels. 


 \begin{figure*}[tb]
    \centering
    \includegraphics[width=1\linewidth]{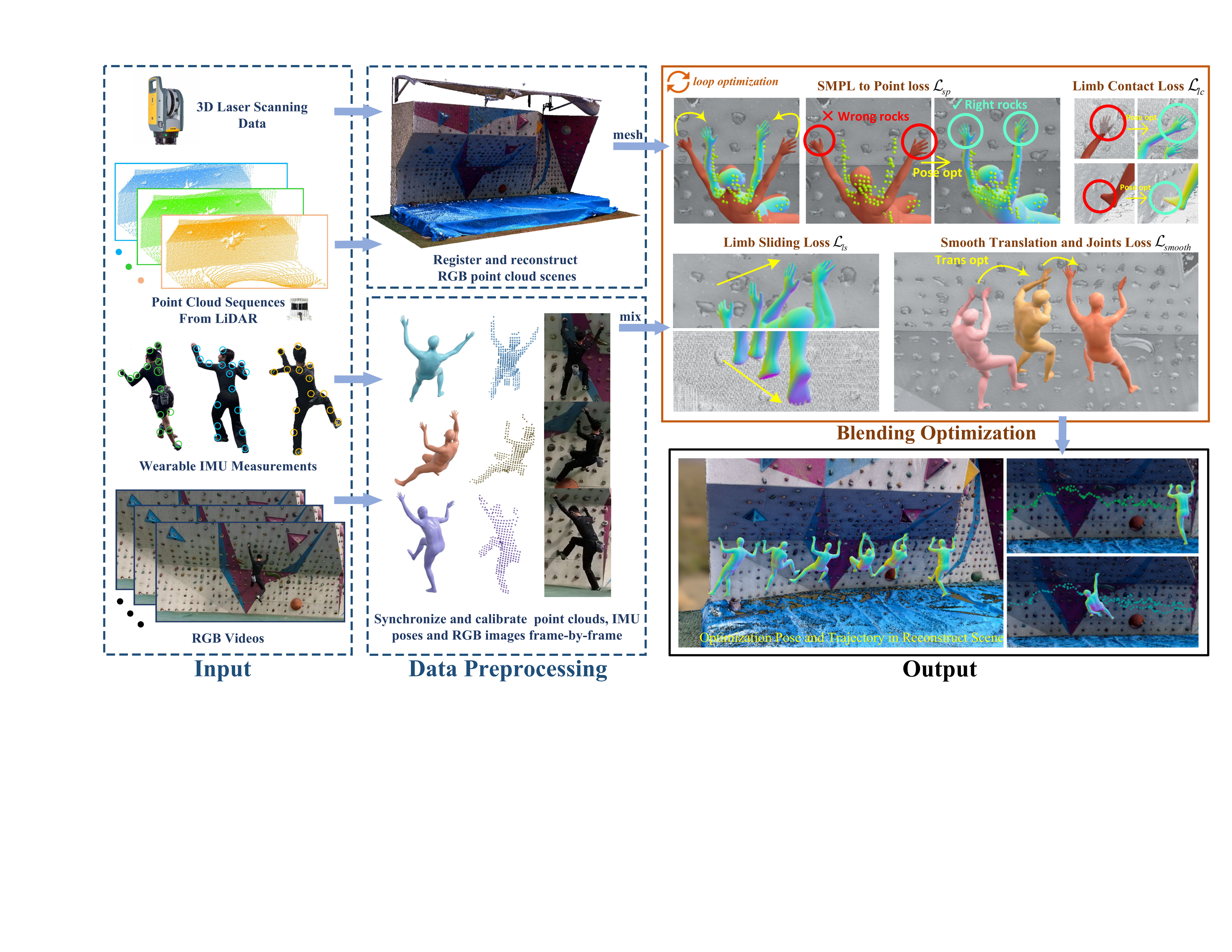}
    \vspace{-6mm}
    \caption{\textbf{Overview of main annotation pipeline.} The blue arrows indicate data flows, and the yellow arrows represent the direction of optimization. \textbf{Dotted box:} The input of each scene consists of RGB videos, point cloud sequence, IMU measurements, and 3D laser scanning data. Data pre-processing stage calibrates and synchronizes different modalities.
     \textbf{Solid box:} The blending optimization stage optimize including the pose and translation based on multiple constraint losses.}
    \vspace{-5mm}
    \label{fig:dataPipeline}
 \end{figure*}

\subsection{Data Annotation Pipeline}
\label{sec:DataPipeline}

The data annotation pipeline consists of 3 stages: preprocessing, blending optimization, and manual annotation. Fig.~\ref{fig:dataPipeline} depicts the preprocessing and the blending optimization stages of the annotation pipeline. \cref{Sec:Data Preprocessing} describe the data preprocessing stage which calibrates and synchronizes multi-modal data. \cref{Sec:Blending optimization} describes the blending optimization stage which uses multiple constraints to improve the quality of human pose and global translation. ~\cref{Sec:Contact Annotation} describes the manual annotation stage.


\subsection{Multi-modal Data Preprocessing Stage}
\label{Sec:Data Preprocessing}




First, we convert high-precision 3D laser scanning data into colored point cloud scenes, followed by conversion of point cloud sequences recorded by the LiDAR into dynamic scenes and register the static and dynamic scenes.  Second, we segment human body point clouds from each frame to assist annotation process, and obtain human pose $\theta$ based on the SMPL~\cite{SMPL2015} model by IMU measurements. Finally, we perform frame-level time synchronization and orientation calibration on the scenes, human poses, human point clouds, and RGB videos.


\PAR{Time synchronization.}
The synchronization between the IMUs, LiDAR, and RGB video is achieved by detecting the peak of a jumping event. In each motion sequence, the actor jumps in place, and we design a peak detection algorithm to find the height peaks in both the IMU's and LiDAR's trajectories automatically. The RGB video and IMU data are down-sampled to 20 FPS, which is consistent with the frame rate of LiDAR. Finally, the LiDAR, RGB video and IMU are synchronized based on the timestamp of the peak.



\PAR{Pose and Translation Initialization} 
A person's motion sequence in world coordinate \{$W$\} is denoted by $\bm{M^W}= (T^W, \theta^W, \beta)$. The $T^I$ and $\theta^I$ in $M^I = (T^I, \theta^I, \beta)$ are provided by the MoCap devices. We use $\theta^W =R_{WI} \theta^I$ to the pose, where $R_{WI}$ is the coarse calibration matrix from $I$ to $W$, and compute the center of gravity of the human body point cloud as the initial translation.





\subsection{Blending Optimization Stage}
\label{Sec:Blending optimization}


We utilize scene and physical constraints to perform a blending optimization of pose and translation to obtain accurate and scene-natural human motion $\bm{M^W}$ annotation. The following constraints are used: the limb contact constraint $\mathcal{L}_{lc}$ encourages reasonable hand and foot contact with the scene mesh without penetrating. The limb sliding constraint $\mathcal{L}_{ls}$ eliminates the unreasonable slippage of the limbs during climbing. The smoothness constraint $\mathcal{L}_{smooth}$ makes the translation, orientation, and joints remain temporal continuity. The SMPL to point constraints $\mathcal{L}_{sp}$ minimizing the distance between constructed SMPL vertices to the point clouds of human body. Please refer to the supplementary material for detailed formulation of the constraints. 

The optimization is expressed as:
\begin{equation}
	\begin{split}
    \mathcal{L}=
        \lambda_{lc}\mathcal{L}_{lc} &+
        \lambda_{ls}\mathcal{L}_{ls} +
        \mathcal{L}_{smooth} +\lambda_{sp}\mathcal{L}_{sp}
    \end{split}
\end{equation}
\noindent
where $\lambda_{lc},\ \lambda_{ls},\ \lambda_{sp}$ are coefficients of loss terms. $\mathcal{L}$ is minimized with a gradient descent algorithm that optimize $\bm{M^W}=(T, \theta)$. $\bm{M^W}$ is initialized in \cref{Sec:Data Preprocessing}.

\PAR{Limb contact Loss.} This loss is defined as the distance between a stable foot or hand and its neighbor among the scene vertices.
First, we detect the state of the foot and hand based on their movements, which are calculated using the set of vertices of hands and feet. One limb is marked as stable if its movement is smaller than 3$cm$ and smaller than another limb (foot or hand)'s movement. Subsequently, we perform a neighbor search to obtain the contact environment in the vicinity of the stable limb. The limb contact loss is $\mathcal{L}_{lc} = \mathcal{L}_{\text {lc}_{feet}} + \mathcal{L}_{\text {lc}_{hand}}$.


\PAR{Limb sliding Loss.} This loss reduces the motion's sliding on the contact surfaces, making the motion more natural and smooth. The sliding loss is defined as the distance of a stable limb over every two successive frames: $\mathcal{L}_{ls} = \mathcal{L}_{\text {ls}_{feet}} + \mathcal{L}_{\text {ls}_{hands}}$.


\PAR{Smooth Loss.} The smooth loss includes the translation term $\mathcal{L}_{trans}$ and the joints term $\mathcal{L}_{joints}$. 
\begin{equation}
	\begin{split}
        \mathcal{L}_{smooth} = 
        \lambda_{trans} \mathcal{L}_{trans} + 
        \lambda_{joints} \mathcal{L}_{joints}
    \end{split}
\end{equation}
The $\mathcal{L}_{trans}$ smooths the trajectory $T$ of human (the translation of the pelvis) through minimizing the difference between LiDAR and a human's translation difference.
The $\mathcal{L}_{joints}$ is the term that smooths the motion of body joints in global 3D space, which minimizes the mean acceleration of the joints. For this loss, we only consider stable joints in the trunk and neck regions. $\lambda_{trans},\ \lambda_{joints}$ are coefficients.

\PAR{SMPL to point loss.} %
For each estimated human meshes, we use Hidden Points Removal (HPR) \cite{katz2007direct} to remove the invisible mesh vertices from the perspective of LiDAR. Then, we use Iterative closest point (ICP)~\cite{segal2009generalized} to register the visible vertices to $\mathcal{P}$, which is segmented human point clouds. We re-project the human body mesh in the LiDAR coordinate to select the visible human body vertices $V'\,\!$. For each frame, We use $\mathcal{L}_{sp}$ to minimize the 3D Chamfer distance between human points $\mathcal{P}_{i}$ and vertices $V'\,\!_{i}$. More details about loss terms definition are given in the appendix.

\subsection{Manual Annotation Stage}
\label{Sec:Contact Annotation}

\PAR{Pose and translation annotation.} After the optimization stage, the human poses and translations are mostly well aligned. For some artifacts, we manually change the pose and the translation parameters of a climber's motions. 
\PAR{Scene contact annotation.} When climbing on a rock wall, a person should apply physical forces on rock holds to climb up. For an in-depth human-scene understanding of the climbing activities, we annotate all the climbing holds in the scene. Further, we have annotated the hands/feet when they contact with the holds for some motion sequences. 
\PAR{Cross verification.} We have invited two external researchers to inspect our dataset. And we have manually corrected the artifacts discovered by them.



\label{sec:ConstructingData}

\section{Dataset Evaluations}

In this section, the CIMI4D dataset quality is demonstrated through qualitative and quantitative evaluations.

\begin{figure}[!tb]
    \vspace{-1mm}
    \centering
     \includegraphics[width=1\linewidth]{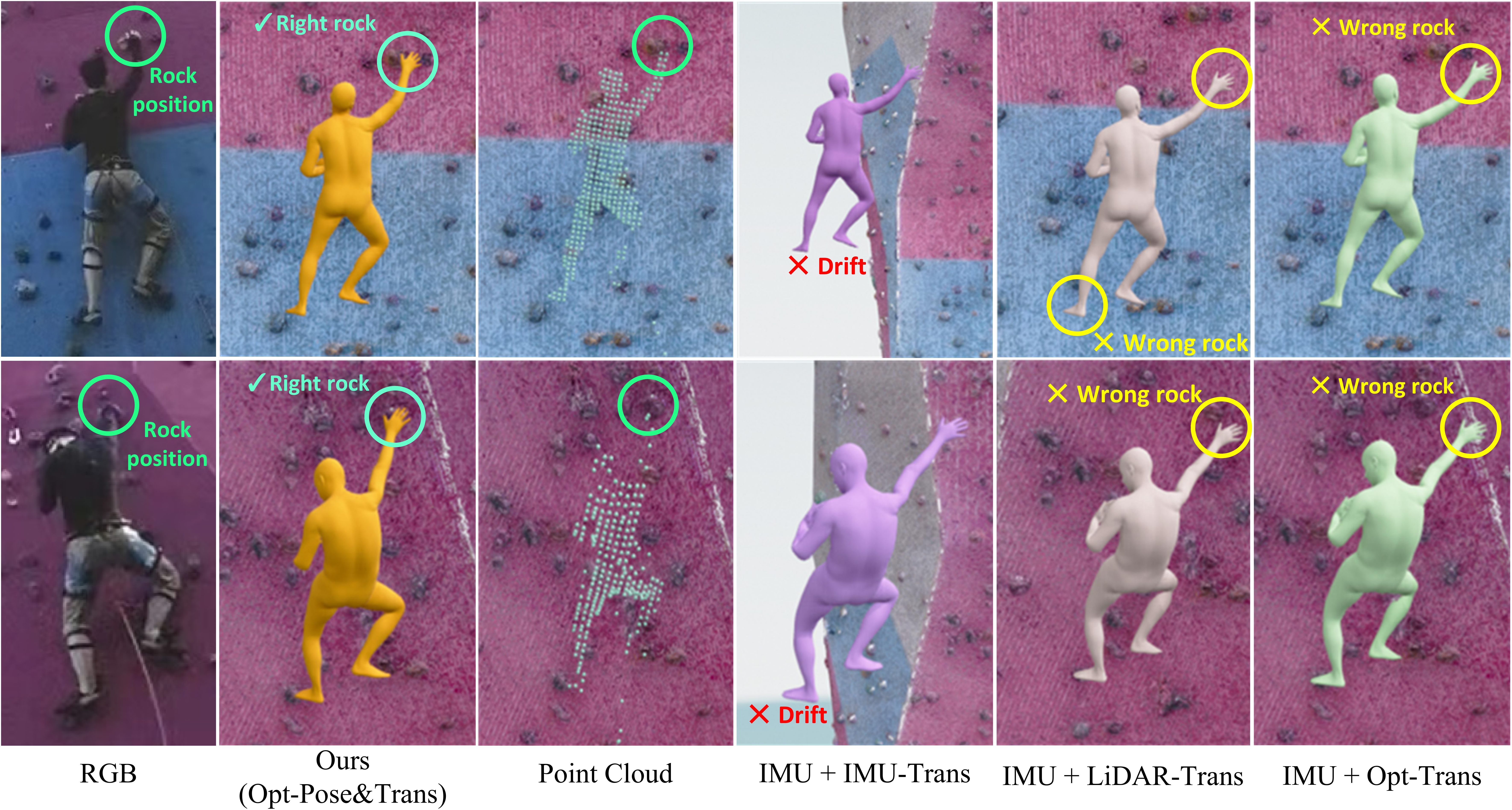}
     \vspace{-6mm}
     \caption{\textbf{Qualitative evaluation.} From left to right: RGB image, our method with the preprocessing and optimization stage, LiDAR point clouds, IMU pose and translation, after preprocessing stage (IMU+LiDAR Trans), optimization stage without the smooth loss (IMU+Opt-Trans).}
     \label{fig:QualitativeCIMI}
     \vspace{-1mm}
\end{figure}

\begin{table}[!tb]
	\centering
	\resizebox{\linewidth}{!}{
    \begin{tabular}{cccccc}
     \toprule
     Scene & ACCEL$\downarrow$ & PMPJPE$\downarrow$ & MPJPE$\downarrow$ &  PVE$\downarrow$ & PCK0.5$\uparrow$\\
     \midrule
      Vertical\;1 & 0.57 & 2.04 & 6.52 & 8.08 & 0.99\\
      Horizontal\;1 & 0.50 & 1.96 & 4.26 & 6.27 & 0.99\\
     \bottomrule
     \end{tabular}%
     }
    \vspace{-3mm}
    \caption{Quantitative evaluation of annotations for two scenes.}
    \vspace{-4mm}
	\label{tab:eva_annotation}
\end{table}

\begin{table}[!tb]
    \centering
	\resizebox{\linewidth}{!}{
    \begin{tabular}{ccc|cccc}
    \toprule
    \multicolumn{3}{c|}{Constraint term} & \multicolumn{4}{c}{Scene} \\
    \midrule
    $\mathcal{L}_{lc}$ & $\mathcal{L}_{smooth}$ & $\mathcal{L}_{sp}$ & Vertical $1$ & Vertical $2$ & Horizontal $1$  & Horizontal $2$   \\
    \midrule
    \textcolor{red}{\XSolidBrush} & \textcolor{red}{\XSolidBrush} &  \textcolor{red}{\XSolidBrush} & 48.28 & 60.04 & 59.83  & 47.74  \\
    \textcolor{green}{\Checkmark} & \textcolor{green}{\Checkmark} &  \textcolor{red}{\XSolidBrush} & 22.64 & 28.33 & 41.67  & 26.64  \\
    \textcolor{green}{\Checkmark} & \textcolor{red}{\XSolidBrush} &  \textcolor{green}{\Checkmark} & 33.48 & 40.44 &  44.77   & 31.44   \\
    \textcolor{red}{\XSolidBrush} & \textcolor{green}{\Checkmark} &  \textcolor{green}{\Checkmark} & 24.64 & 38.37 &  42.07  & 30.08  \\
    \midrule
    \textcolor{green}{\Checkmark} & \textcolor{green}{\Checkmark} &  \textcolor{green}{\Checkmark} & \textbf{16.24} & \textbf{23.46} & \textbf{34.34}   & \textbf{20.21}  \\
    \bottomrule
    \end{tabular}%
    }
    \vspace{-3mm}
    \caption{Loss of the optimization stage for different constraints}
    \vspace{-5mm}
    \label{tab:eva_CIMI}
\end{table}

\PAR{Qualitative comparison.}\cref{fig:QualitativeCIMI} depicts a frame of the CIMI4D dataset. As it is shown in the figure, the LiDAR can obtain the point clouds of a human, but it does not contain the translation and pose of a human. The IMU poses (IMU+IMU-trans) drift over time, its translation is not correct. The preprocessing stage (IMU+LiDAR-Trans) does improve the quality of the data. However, IMUs are impacted by the magnetic field of the wall, which contains rebars. IMU pose mistakenly touches a wrong rock point. Using the optimization stage without the smooth loss (IMU+Opt-Trans) improves the quality of annotations with fewer number of wrong touch than IMU+LiDAR-Trans. Our method can accurately reconstruct the pose and translation of a person.

\begin{figure*}[!tb]
    \vspace{-3mm}
    \centering
     \includegraphics[width=1\linewidth]{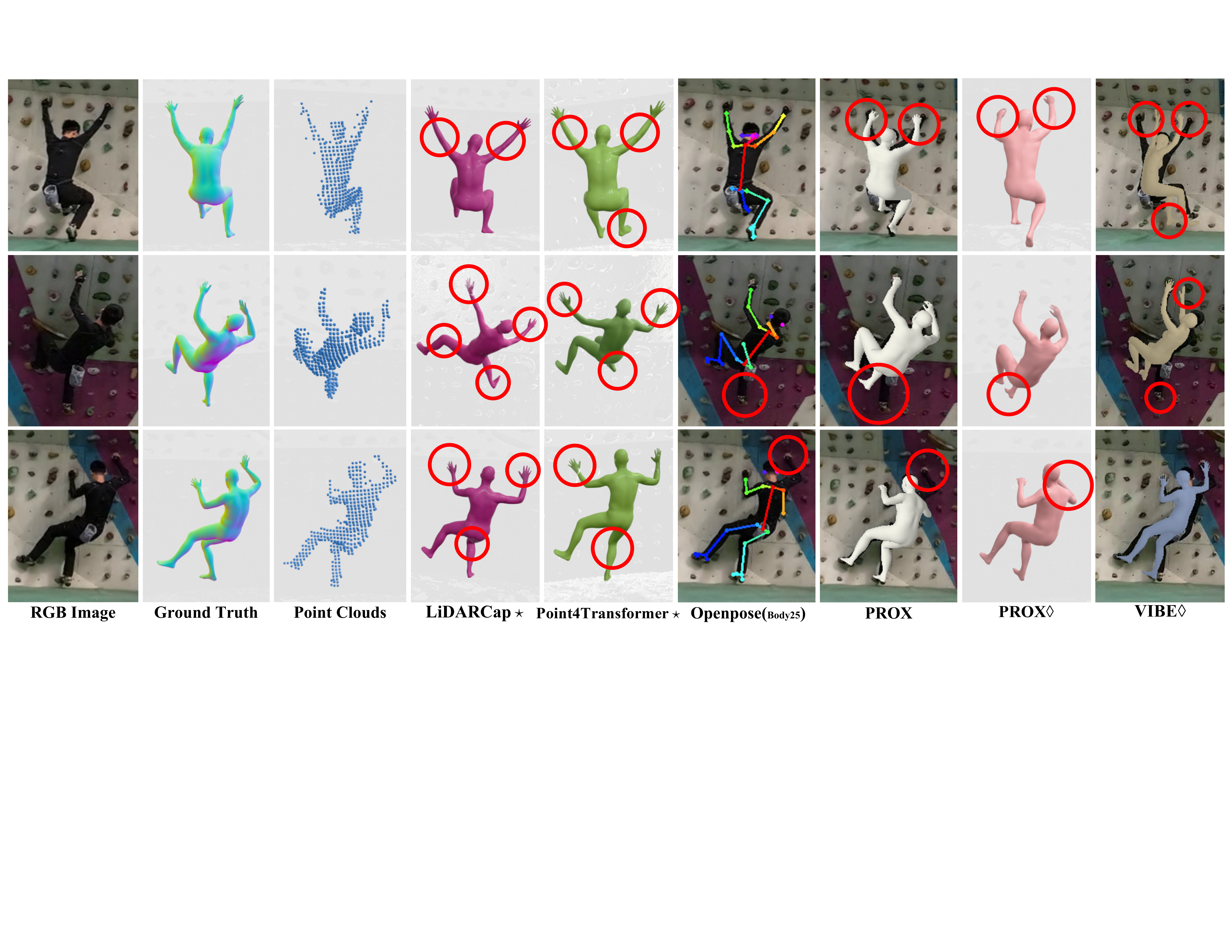}
     \vspace{-6mm}
     \caption{Qualitative results of several algorithms on the CIMI4D dataset. It is challenging to reconstruct a climbing pose with high ductility, even if algorithms are re-trained (marked by $\star$) or fine-tuned (marked by $\diamond$) based on CIMI4D. As indicated by the red circles, all these methods have artifacts for limbs. As a scene-aware method, PROX performs better than other methods that do not use scene constraints. This suggests that it is necessary to consider the human-scene interaction annotation provided in CIMI4D.}
    \vspace{-3mm}
     \label{fig:QualitativeTask}
\end{figure*}

\PAR{Evaluation metrics.} In this section and in ~\cref{sec:tasks}, we report Procrustes-Aligned Mean Per Joint Position Error (PMPJPE), Mean Per Joint Position Error (MPJPE), Percentage of Correct Keypoints (PCK), Per Vertex Error (PVE), and Acceleration error($m/s^2$) (ACCEL). Except ACCEL, error metrics are measured in millimeters.


\PAR{Quantitative evaluation.} To quantitatively evaluate the annotation quality of CIMI4D, we have manually annotated motion sequences from two scenes. And then evaluate the performance of the optimization stage by comparing the annotations generated by the optimization stage (in~\cref{Sec:Blending optimization}) with manual annotations. 

\cref{tab:eva_annotation} shows the error metrics of the annotations generated by the optimization stage, the errors are quite small. This indicates that the effectiveness of the annotation pipeline, and suggests that the high quality of CIMI4D.

Further, to understand the impact of different constraints used in the optimization stage, we conduct ablation study of 3 different constraints: $\mathcal{L}_{cont}$, $\mathcal{L}_{smt}$ and $\mathcal{L}_{stp}$. ~\cref{tab:eva_CIMI} shows the loss of using different combinations of constraints for motions from 4 scenes. The loss is an indicator of violation of motion constraints. Without using any term, the loss is largest, which suggests that motions may seem unnatural. The $\mathcal{L}_{ct}$ and $\mathcal{L}_{smt}$ terms can reduce total loss, which indicates that they can improve the overall quality of data. Combing $\mathcal{L}_{stp}$ can further improve the quality of motions.  Overall, all constraint terms are necessary to produce accurate and smooth human pose and translation. 


\section{Tasks and Benchmarks}\label{sec:tasks}

In this section, we perform an in-depth analysis on the performance of state-of-the-art approaches on the CIMI4D dataset. To evaluate the merit of the CIMI4D dataset, we consider four tasks: 3D pose estimation, 3D pose estimation with scene constraints, motion prediction with scene constraints, and motion generation with scene constraints. We have randomly split the motion sequences with a ratio of 7:3 into training and test sets. We provide results of baseline methods and existing methods. The experimental results for these tasks and the new challenge brought by CIMI4D are discussed in this section.



\subsection{3D Pose Estimation}

\PAR{Pose Estimation.} In this task, the poses of climbing humans are estimated from RGB imagery or LiDAR point clouds based on the CIMI4D dataset. For the methods evaluated in this section, VIBE~\cite{VIBE}, MEAD~\cite{MAED} and DynaBOA~\cite{DynaBOA} estimate poses from RGB images, while LiDARCap and P4Transformer~\cite{Fan2021Point4T} recover the motions from point clouds. The qualitative results of pose estimation are depicted in \cref{fig:QualitativeTask}. As it is pointed out by the red circles in this figure, all these methods have artifacts. 
The quantitative results are depicted in \cref{tab:single_pose_estimation}. The pretrained LiDARCap model performs bad (PCK0.5$=0.46$) on CIMI4D. Further, we train LiDARCap and P4Transformer on CIMI4D. 
The RGB-based approach (VIBE) does not perform good on this dataset too. After fine-tuning on CIMI4D, the performance of VIBE is improved. However, the performance is still poor compared to the original paper. 
Overall, the error metrics for all these methods are increasing, which indicates that CIMI4D is a challenging dataset for pose estimation. 



\begin{table}[tb]
	\centering

	\centering
	\resizebox{\linewidth}{!}{
    \begin{tabular}{ccccccc}
     \toprule
     Input & Method & ACCEL$\downarrow$ & PMPJPE$\downarrow$ & MPJPE$\downarrow$ & PVE$\downarrow$ & PCK0.5$\uparrow$ \\
     \midrule
     \multirow{3}*{LiDAR}  & LiDARCap   & 12.39 & 222.11 & 358.13 & 422.65 & 0.50 \\
        ~ & LiDARCap$\star$   & 2.59 & 86.38 & 115.93 & 136.83 & 0.90 \\
        ~ & P4Transformer$\star$ & 3.32 & 100.58 & 130.99 & 156.27 & 0.87 \\
     \midrule
    \multirow{4}*{RGB} & VIBE  &  68.02 & 287.14 & 770.77 & 857.83 & 0.17  \\
        ~ & VIBE$\diamond$  & 57.88 & 116.78 & 161.21 & 187.70 & 0.76 \\
        ~ & MAED$\diamond$  & 17.50 & 135.57 & 170.43 & 197.66 & 0.74 \\
        ~ & DynaBOA  &  52.4 & 230.86 & 303.16  &  285.62 & 0.54 \\
    \midrule
    \multirow{3}*{Scene} & PROX & - & 109.34 & 265.34 & 279.50 & 0.53 \\
        & PROX$\diamond$ & - & 109.33 & 147.41 & 165.12 & 0.79\\
        & LEMO  & 98.3 & 317.64 & 669.38 & 359.11 & 0.45 \\
     \bottomrule
     \end{tabular}%
     }
     \vspace{-3mm}
     \caption{Comparison of pose estimation by SOTA on different modal data. $\star$ indicates training based on the CIMI4D dataset. $\diamond$ denotes fine-tuned based on the CIMI4D dataset. Other experiments used the pretrained model of the original method.}
     \vspace{-5mm}
	\label{tab:single_pose_estimation}
\end{table}



\PAR{Pose Estimation with Scene Constraints.}
PROX~\cite{PROX} and LEMO~\cite{LEMO} are common-used pose estimation method with scene constraints. To test them on CIMI4D, we obtain skeleton information from OpenPose\cite{OpenPose}, and convert the scene of CIMI4D into \textit{sdf} form to build as the inputs for them. As shown in~\cref{tab:single_pose_estimation}, PROX has large estimation error on CIMI4D. Further, we fine-tune PROX on CIMI4D. Albeit its performance improves, the algorithm should be further improved to obtain satisfactory performance.    

\cref{fig:QualitativeTask} depicts the qualitative results for PROX and LEMO. They rely upon others to provide 2D skeleton information. For such challenging poses with self-occlusion and color similarity among humans and scene, 2D method (i.e., OpenPose) fails. The human joints reconstructed by PROX and LEMO have serious deviations, and the movements of the volunteers are not correctly restored.

\begin{figure}[!tb]
    \vspace{-2mm}
    \centering
     \includegraphics[width=0.98\linewidth]{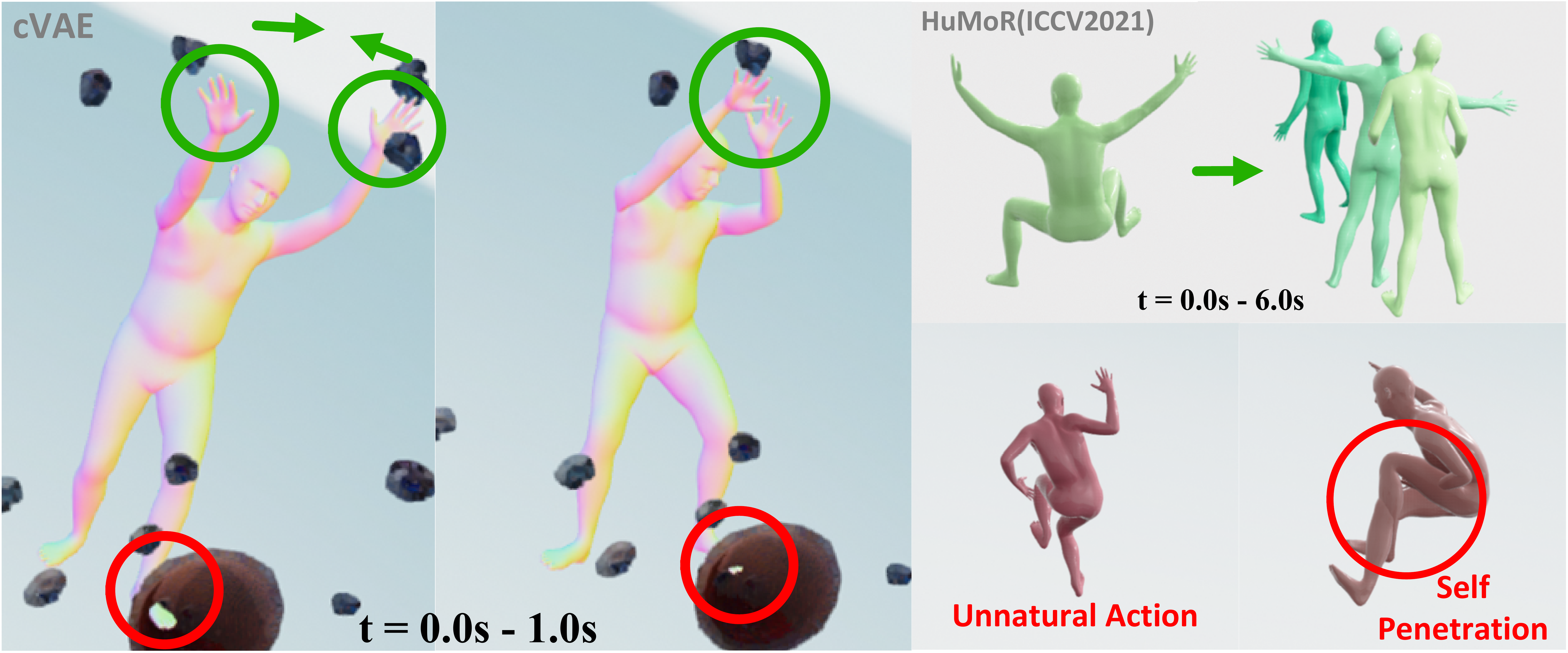}
     \vspace{-2.5mm}
     \caption{
     Predicted body pose and translation. The gesture range predicted by cVEA is slight, but reasonable. HuMor cannot predict climbing motions. The prior dataset AMASS make HuMor's movements eventually become daily motions, and generate many unnatural and self-penetrating actions.
     }
    \vspace{-6mm} 
     \label{fig:QualitativeCIMI6}
\end{figure}
\subsection{Motion Prediction and Generation}

The rich annotations of CIMI4D enable us to explore new motion-related tasks. The two tasks explored in this section are motion prediction and motion generation tasks with scene constraints. 

\PAR{Motion Prediction.}
In this task, we predict the pose and the global trajectory of a person in the future based on one previous frame. This task is more difficult than merely predicting poses. As it is shown in \cref{tab:single_pose_estimation}, LiDAR-based approaches perform better than RGB-based approaches. Thus, we predict motions based on point clouds. 

We design a straightforward baseline architecture that utilizes LiDARCap~\cite{lidarcap} as its backbone and incorporates a conditional variational autoencoder to capture the distribution of human motion.
We also input climbing body keyframes to HuMor~\cite{HUMOR_ICCV2021}. They predict the poses and translations of a human in the next few second.  ~\cref{fig:QualitativeCIMI6} shows the prediction results, although there are translation errors and penetration artifacts, we can observe natural and smooth hand movements, as indicated by the green arrow in cVAE. Since HuMor uses AMASS as a priori action, it cannot predict climbing behavior, and all climbing actions eventually become other motions. Many unreasonable movements can be seen in the picture. It is challenging to predict motions using the CIMI4D dataset.



\PAR{Motion Generation with Scene Constraints.}
Given a scene, it is interesting to generate a physically plausible pose for better human-scene understanding. For example, it is important for climbers to estimate possible poses for a specific set of rock holds thus to climb up. 
For this tasks, we design a baseline which uses a conditional variational autoencoder model to generate physically plausible pose. To test the baseline, we choose some rock holds that model has not seen before and then generates poses and translations with physical plausibility.

\begin{figure}[!tb]
    \centering
     \includegraphics[width=1\linewidth]{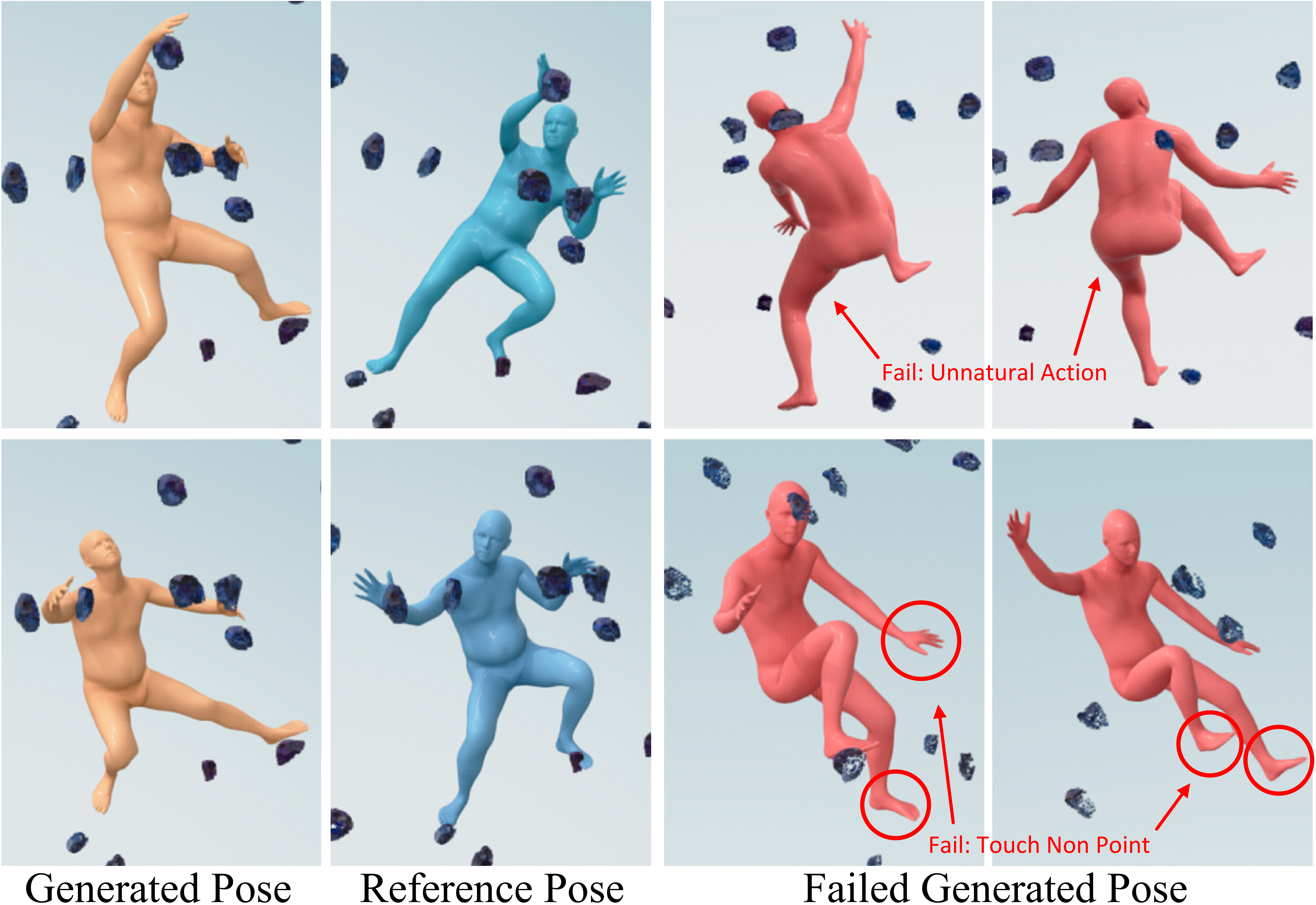}
     \vspace{-6mm}
     \caption{Generated climbing poses with unseen holds. Blue poses are reference poses. Orange and red poses are generated by the baseline. Professional climber consider these orange poses to be reasonable and natural, and the red generated poses are unrealistic.}
    \vspace{-5mm}
     \label{fig:QualitativeCIMI7}
\end{figure}

\cref{fig:QualitativeCIMI7} depicts examples of generated poses and translations. For some sets of holds, it is possible to generate reasonable poses. But for some other sets of holds, the baseline fails. Overall, the diversity of the motion generation algorithm is small. It is challenging to generate poses and translations with scene constraints. For more details on these two tasks, please refer to the supplementary materials.




\section{Limitations and Future Work}
\vspace{-2mm}
There are three major limitations of CIMI4D. Firstly, CIMI4D does not contain detailed hand poses, it leads to a slight penetrating with the holds. This can be addressed in future work by using MoCap gloves with the SMPL-X model. Secondly, CIMI4D record the poses of climbers, but it does not contain a fine category-level annotation of the climbing actions. A fine-grain annotation of climbing motions could enrich this community better. Thirdly, we focus on the reconstruction of human motions whereas ignoring the photo-realistic reconstruction of 3D scenes. Using neural rendering techniques~\cite{nerf} to reconstruct 3D scenes and humans may worth exploring.



\vspace{-2mm}
\section{Conclusion}
\vspace{-2mm}
We propose CIMI4D, the first 3D climbing dataset with complex movements and scenes. CIMI4D consists of 180K frames of RGB videos, LiDAR point clouds, IMU measurements with precise annotations, and 13 high-precision scenes. We annotate the dataset more accurately by blending optimization. Besides human pose estimation tasks, the rich annotations in CIMI4D enable benchmarking on scene-aware tasks such as motion prediction and motion generation.  We evaluate multiple methods for these tasks, and the results demonstrate that CIMI4D presents new challenges to today’s computer vision approaches.
\vspace{+1mm}
\PAR{Acknowledgement.}This work was partially supported by the open fund of PDL (2022-KJWPDL-12, WDZC20215250113), the FuXiaQuan National Independent Innovation Demonstration Zone Collaborative Innovation Platform(No.3502ZCQXT2021003), and the Fundamental Research
Funds for the Central Universities (No.20720220064). We thank Li Men and Peifang Xu for data collection and advice. We thank Yan Zhang, Shuqiang Cai and Minghang Zhu for paper checking.



{\small
\bibliographystyle{ieee_fullname}
\bibliography{egbib}

\begin{thebibliography}{10}\itemsep=-1pt

\bibitem{alldieck2017optical}
Thiemo Alldieck, Marc Kassubeck, Bastian Wandt, Bodo Rosenhahn, and Marcus
  Magnor.
\newblock Optical flow-based 3d human motion estimation from monocular video.
\newblock In {\em German Conference on Pattern Recognition}, pages 347--360.
  Springer, 2017.

\bibitem{ClimbingSensorSurvey2022}
Marina Andric, Francesco Ricci, and Floriano Zini.
\newblock Sensor-based activity recognition and performance assessment in
  climbing: A review.
\newblock {\em IEEE Access}, 10:108583--108603, 2022.

\bibitem{Andriluka2018PoseTrackAB}
Mykhaylo Andriluka, Umar Iqbal, Anton Milan, Eldar Insafutdinov, Leonid
  Pishchulin, Juergen Gall, and Bernt Schiele.
\newblock Posetrack: A benchmark for human pose estimation and tracking.
\newblock {\em 2018 IEEE/CVF Conference on Computer Vision and Pattern
  Recognition}, pages 5167--5176, 2018.

\bibitem{Beltrn2022AutomatedHM}
Raul~Beltr{\'a}n Beltr{\'a}n, Julia Richter, and Ulrich Heinkel.
\newblock Automated human movement segmentation by means of human pose
  estimation in rgb-d videos for climbing motion analysis.
\newblock In {\em VISIGRAPP}, 2022.

\bibitem{bhatnagar2022behave}
Bharat~Lal Bhatnagar, Xianghui Xie, Ilya~A Petrov, Cristian Sminchisescu,
  Christian Theobalt, and Gerard Pons-Moll.
\newblock Behave: Dataset and method for tracking human object interactions.
\newblock In {\em Proceedings of the IEEE/CVF Conference on Computer Vision and
  Pattern Recognition}, pages 15935--15946, 2022.

\bibitem{Bregler1998TrackingPW}
Christoph Bregler and Jitendra Malik.
\newblock Tracking people with twists and exponential maps.
\newblock {\em Proceedings. 1998 IEEE Computer Society Conference on Computer
  Vision and Pattern Recognition (Cat. No.98CB36231)}, pages 8--15, 1998.

\bibitem{OpenPose}
Zhe Cao, Tomas Simon, Shih-En Wei, and Yaser Sheikh.
\newblock Realtime multi-person 2d pose estimation using part affinity fields.
\newblock In {\em Computer Vision and Pattern Recognition (CVPR)}, 2017.

\bibitem{Carreira2017QuoVA}
Jo{\~a}o Carreira and Andrew Zisserman.
\newblock Quo vadis, action recognition? a new model and the kinetics dataset.
\newblock {\em 2017 IEEE Conference on Computer Vision and Pattern Recognition
  (CVPR)}, pages 4724--4733, 2017.

\bibitem{HSC4D}
Yudi Dai, Yitai Lin, Chenglu Wen, Siqi Shen, Lan Xu, Jingyi Yu, Yuexin Ma, and
  Cheng Wang.
\newblock Hsc4d: Human-centered 4d scene capture in large-scale indoor-outdoor
  space using wearable imus and lidar.
\newblock In {\em Proceedings of the IEEE/CVF Conference on Computer Vision and
  Pattern Recognition (CVPR)}, pages 6792--6802, June 2022.

\bibitem{dou-siggraph2016}
Mingsong Dou, Sameh Khamis, Yury Degtyarev, Philip Davidson, Sean Fanello,
  Adarsh Kowdle, Sergio~Orts Escolano, Christoph Rhemann, David Kim, Jonathan
  Taylor, Pushmeet Kohli, Vladimir Tankovich, and Shahram Izadi.
\newblock {Fusion4D: Real-time Performance Capture of Challenging Scenes}.
\newblock In {\em ACM SIGGRAPH Conference on Computer Graphics and Interactive
  Techniques}, 2016.

\bibitem{elias2021speed21}
Petr Elias, Veronika Skvarlova, and Pavel Zezula.
\newblock Speed21: Speed climbing motion dataset.
\newblock In {\em Proceedings of the 4th International Workshop on Multimedia
  Content Analysis in Sports}, pages 43--50, 2021.

\bibitem{Fan2021Point4T}
Hehe Fan, Yi Yang, and Mohan~S. Kankanhalli.
\newblock Point 4d transformer networks for spatio-temporal modeling in point
  cloud videos.
\newblock {\em 2021 IEEE/CVF Conference on Computer Vision and Pattern
  Recognition (CVPR)}, pages 14199--14208, 2021.

\bibitem{DynaBOA}
Shanyan Guan, Jingwei Xu, Michelle~Z He, Yunbo Wang, Bingbing Ni, and Xiaokang
  Yang.
\newblock Out-of-domain human mesh reconstruction via dynamic bilevel online
  adaptation.
\newblock {\em IEEE Transactions on Pattern Analysis and Machine Intelligence},
  2022.

\bibitem{HPS}
Vladimir Guzov, Aymen Mir, Torsten Sattler, and Gerard Pons-Moll.
\newblock Human poseitioning system (hps): 3d human pose estimation and
  self-localization in large scenes from body-mounted sensors.
\newblock In {\em Proceedings of the IEEE/CVF Conference on Computer Vision and
  Pattern Recognition}, pages 4318--4329, 2021.

\bibitem{LiveCap2019tog}
Marc Habermann, Weipeng Xu, Michael Zollh\"{o}fer, Gerard Pons-Moll, and
  Christian Theobalt.
\newblock Livecap: Real-time human performance capture from monocular video.
\newblock {\em ACM Transactions on Graphics (TOG)}, 38(2):14:1--14:17, 2019.

\bibitem{DeepCap_CVPR2020}
Marc Habermann, Weipeng Xu, Michael Zollhofer, Gerard Pons-Moll, and Christian
  Theobalt.
\newblock Deepcap: Monocular human performance capture using weak supervision.
\newblock In {\em Proceedings of the IEEE/CVF Conference on Computer Vision and
  Pattern Recognition (CVPR)}, June 2020.

\bibitem{PROX}
Mohamed Hassan, Vasileios Choutas, Dimitrios Tzionas, and Michael~J. Black.
\newblock Resolving 3d human pose ambiguities with 3d scene constraints.
\newblock In {\em 2019 {IEEE/CVF} International Conference on Computer Vision,
  {ICCV} 2019, Seoul, Korea (South), October 27 - November 2, 2019}, pages
  2282--2292. {IEEE}, 2019.

\bibitem{POSER}
Mohamed Hassan, Partha Ghosh, Joachim Tesch, Dimitrios Tzionas, and Michael~J.
  Black.
\newblock Populating 3d scenes by learning human-scene interaction.
\newblock In {\em {IEEE} Conference on Computer Vision and Pattern Recognition,
  {CVPR} 2021, virtual, June 19-25, 2021}, pages 14708--14718. Computer Vision
  Foundation / {IEEE}, 2021.

\bibitem{challencap}
Yannan He, Anqi Pang, Xin Chen, Han Liang, Minye Wu, Yuexin Ma, and Lan Xu.
\newblock Challencap: Monocular 3d capture of challenging human performances
  using multi-modal references.
\newblock In {\em Proceedings of the IEEE/CVF Conference on Computer Vision and
  Pattern Recognition}, pages 11400--11411, 2021.

\bibitem{RICH}
Chun-Hao~P Huang, Hongwei Yi, Markus H{\"o}schle, Matvey Safroshkin, Tsvetelina
  Alexiadis, Senya Polikovsky, Daniel Scharstein, and Michael~J Black.
\newblock Capturing and inferring dense full-body human-scene contact.
\newblock In {\em Proceedings of the IEEE/CVF Conference on Computer Vision and
  Pattern Recognition}, pages 13274--13285, 2022.

\bibitem{DIP}
Yinghao Huang, Manuel Kaufmann, Emre Aksan, Michael~J. Black, Otmar Hilliges,
  and Gerard Pons-Moll.
\newblock Deep inertial poser: Learning to reconstruct human pose from sparse
  inertial measurements in real time.
\newblock {\em ACM Transactions on Graphics, (Proc. SIGGRAPH Asia)},
  37(6):185:1--185:15, nov 2018.

\bibitem{ViconForceClimbing}
Hitomi Iguma, Akihiro Kawamura, and Ryo Kurazume.
\newblock A new 3d motion and force measurement system for sport climbing.
\newblock In {\em 2020 {IEEE/SICE} International Symposium on System
  Integration, {SII} 2020, Honolulu, HI, USA, January 12-15, 2020}, pages
  1002--1007. {IEEE}, 2020.

\bibitem{Ionescu2014Human36MLS}
Catalin Ionescu, Dragos Papava, Vlad Olaru, and Cristian Sminchisescu.
\newblock Human3.6m: Large scale datasets and predictive methods for 3d human
  sensing in natural environments.
\newblock {\em IEEE Transactions on Pattern Analysis and Machine Intelligence},
  36:1325--1339, 2014.

\bibitem{HMR}
Angjoo Kanazawa, Michael~J. Black, David~W. Jacobs, and Jitendra Malik.
\newblock End-to-end recovery of human shape and pose.
\newblock In {\em Computer Vision and Pattern Regognition (CVPR)}, 2018.

\bibitem{Kanazawa2019Learning3H}
Angjoo Kanazawa, Jason~Y. Zhang, Panna Felsen, and Jitendra Malik.
\newblock Learning 3d human dynamics from video.
\newblock {\em 2019 IEEE/CVF Conference on Computer Vision and Pattern
  Recognition (CVPR)}, pages 5607--5616, 2019.

\bibitem{katz2007direct}
Sagi Katz, Ayellet Tal, and Ronen Basri.
\newblock Direct visibility of point sets.
\newblock In {\em ACM SIGGRAPH 2007 papers}, pages 24--es. 2007.

\bibitem{Kim2019PedXBD}
Wonhui Kim, Manikandasriram~Srinivasan Ramanagopal, Charlie Barto, Ming-Yuan
  Yu, Karl Rosaen, Nicholas Goumas, Ram Vasudevan, and Matthew
  Johnson-Roberson.
\newblock Pedx: Benchmark dataset for metric 3-d pose estimation of pedestrians
  in complex urban intersections.
\newblock {\em IEEE Robotics and Automation Letters}, 4:1940--1947, 2019.

\bibitem{VIBE}
Muhammed Kocabas, Nikos Athanasiou, and Michael~J. Black.
\newblock Vibe: Video inference for human body pose and shape estimation.
\newblock {\em 2020 IEEE/CVF Conference on Computer Vision and Pattern
  Recognition (CVPR)}, pages 5252--5262, 2020.

\bibitem{PARE_ICCV2021}
Muhammed Kocabas, Chun-Hao~P. Huang, Otmar Hilliges, and Michael~J. Black.
\newblock Pare: Part attention regressor for 3d human body estimation.
\newblock In {\em Proceedings of the IEEE/CVF International Conference on
  Computer Vision (ICCV)}, pages 11127--11137, October 2021.

\bibitem{SPIN_ICCV2019}
Nikos Kolotouros, Georgios Pavlakos, Michael~J Black, and Kostas Daniilidis.
\newblock Learning to reconstruct 3d human pose and shape via model-fitting in
  the loop.
\newblock In {\em Proceedings of the IEEE/CVF International Conference on
  Computer Vision}, pages 2252--2261, 2019.

\bibitem{MRClimbing17}
Felix Kosmalla, Andr\'{e} Zenner, Marco Speicher, Florian Daiber, Nico Herbig,
  and Antonio Kr\"{u}ger.
\newblock Exploring rock climbing in mixed reality environments.
\newblock In {\em Proceedings of the 2017 CHI Conference Extended Abstracts on
  Human Factors in Computing Systems}, CHI EA '17, page 1787–1793, New York,
  NY, USA, 2017. Association for Computing Machinery.

\bibitem{footVR}
Felix Kosmalla, Andr{\'{e}} Zenner, Corinna Tasch, Florian Daiber, and Antonio
  Kr{\"{u}}ger.
\newblock The importance of virtual hands and feet for virtual reality
  climbing.
\newblock In {\em Extended Abstracts of the 2020 {CHI} Conference on Human
  Factors in Computing Systems, {CHI} 2020, Honolulu, HI, USA, April 25-30,
  2020}, pages 1--8. {ACM}, 2020.

\bibitem{lidarcap}
Jialian Li, Jingyi Zhang, Zhiyong Wang, Siqi Shen, Chenglu Wen, Yuexin Ma, Lan
  Xu, Jingyi Yu, and Cheng Wang.
\newblock Lidarcap: Long-range marker-less 3d human motion capture with lidar
  point clouds.
\newblock In {\em Proceedings of the IEEE/CVF Conference on Computer Vision and
  Pattern Recognition}, pages 20502--20512, 2022.

\bibitem{MiaoLiu20204DHB}
Miao Liu, Dexin Yang, Yan Zhang, Zhaopeng Cui, James~M. Rehg, and Siyu Tang.
\newblock 4d human body capture from egocentric video via 3d scene grounding.
\newblock {\em International Conference on 3d vision}, 2020.

\bibitem{SMPL2015}
Matthew Loper, Naureen Mahmood, Javier Romero, Gerard Pons-Moll, and Michael~J.
  Black.
\newblock Smpl: A skinned multi-person linear model.
\newblock {\em ACM Trans. Graph.}, 34(6):248:1--248:16, Oct. 2015.

\bibitem{AMASS_ICCV2019}
Naureen Mahmood, Nima Ghorbani, Nikolaus~F. Troje, Gerard Pons-Moll, and
  Michael~J. Black.
\newblock Amass: Archive of motion capture as surface shapes.
\newblock In {\em Proceedings of the IEEE/CVF International Conference on
  Computer Vision (ICCV)}, October 2019.

\bibitem{Martinez17}
Julieta Martinez, Rayat Hossain, Javier Romero, and James~J Little.
\newblock A simple yet effective baseline for 3d human pose estimation.
\newblock In {\em ICCV}, 2017.

\bibitem{Mehta2017Monocular3H}
Dushyant Mehta, Helge Rhodin, Dan Casas, Pascal~V. Fua, Oleksandr Sotnychenko,
  Weipeng Xu, and Christian Theobalt.
\newblock Monocular 3d human pose estimation in the wild using improved cnn
  supervision.
\newblock {\em 2017 International Conference on 3D Vision (3DV)}, pages
  506--516, 2017.

\bibitem{nerf}
Ben Mildenhall, Pratul~P. Srinivasan, Matthew Tancik, Jonathan~T. Barron, Ravi
  Ramamoorthi, and Ren Ng.
\newblock Nerf: Representing scenes as neural radiance fields for view
  synthesis.
\newblock In {\em Computer Vision - {ECCV} 2020 - 16th European Conference,
  Glasgow, UK, August 23-28, 2020, Proceedings, Part {I}}, volume 12346 of {\em
  Lecture Notes in Computer Science}, pages 405--421. Springer, 2020.

\bibitem{pandurevic2022analysis}
Dominik Pandurevic, Pawe{\l} Draga, Alexander Sutor, and Klaus Hochradel.
\newblock Analysis of competition and training videos of speed climbing
  athletes using feature and human body keypoint detection algorithms.
\newblock {\em Sensors}, 22(6):2251, 2022.

\bibitem{Patel2020TailorNetPC}
Chaitanya Patel, Zhouyingcheng Liao, and Gerard Pons-Moll.
\newblock Tailornet: Predicting clothing in 3d as a function of human pose,
  shape and garment style.
\newblock {\em 2020 IEEE/CVF Conference on Computer Vision and Pattern
  Recognition (CVPR)}, pages 7363--7373, 2020.

\bibitem{AGORA}
Priyanka Patel, Chun-Hao~P. Huang, Joachim Tesch, David~T. Hoffmann, Shashank
  Tripathi, and Michael~J. Black.
\newblock {AGORA}: Avatars in geography optimized for regression analysis.
\newblock In {\em Proceedings IEEE/CVF Conf.~on Computer Vision and Pattern
  Recognition ({CVPR})}, June 2021.

\bibitem{VPose}
Georgios Pavlakos, Vasileios Choutas, Nima Ghorbani, Timo Bolkart, Ahmed A.~A.
  Osman, Dimitrios Tzionas, and Michael~J. Black.
\newblock Expressive body capture: 3d hands, face, and body from a single
  image.
\newblock In {\em {IEEE} Conference on Computer Vision and Pattern Recognition,
  {CVPR} 2019, Long Beach, CA, USA, June 16-20, 2019}, pages 10975--10985.
  Computer Vision Foundation / {IEEE}, 2019.

\bibitem{pons2011outdoor}
Gerard Pons-Moll, Andreas Baak, Juergen Gall, Laura Leal-Taixe, Meinard
  Mueller, Hans-Peter Seidel, and Bodo Rosenhahn.
\newblock Outdoor human motion capture using inverse kinematics and von
  mises-fisher sampling.
\newblock In {\em 2011 International Conference on Computer Vision}, pages
  1243--1250. IEEE, 2011.

\bibitem{HUMOR_ICCV2021}
Davis Rempe, Tolga Birdal, Aaron Hertzmann, Jimei Yang, Srinath Sridhar, and
  Leonidas~J. Guibas.
\newblock Humor: 3d human motion model for robust pose estimation.
\newblock In {\em Proceedings of the IEEE/CVF International Conference on
  Computer Vision (ICCV)}, pages 11488--11499, October 2021.

\bibitem{reveret20203d}
Lionel Reveret, Sylvain Chapelle, Franck Quaine, and Pierre Legreneur.
\newblock 3d visualization of body motion in speed climbing.
\newblock {\em Frontiers in Psychology}, 11:2188, 2020.

\bibitem{climbingSurvey2020}
Julia Richter, Raul~Beltr{\'a}n Beltr{\'a}n, Guido K{\"o}stermeyer, and Ulrich
  Heinkel.
\newblock Human climbing and bouldering motion analysis: A survey on sensors,
  motion capture, analysis algorithms, recent advances and applications.
\newblock In {\em VISIGRAPP}, 2020.

\bibitem{pifu}
Shunsuke Saito, Zeng Huang, Ryota Natsume, Shigeo Morishima, Hao Li, and Angjoo
  Kanazawa.
\newblock Pifu: Pixel-aligned implicit function for high-resolution clothed
  human digitization.
\newblock In {\em 2019 {IEEE/CVF} International Conference on Computer Vision,
  {ICCV} 2019, Seoul, Korea (South), October 27 - November 2, 2019}, pages
  2304--2314. {IEEE}, 2019.

\bibitem{pifuhd}
Shunsuke Saito, Tomas Simon, Jason~M. Saragih, and Hanbyul Joo.
\newblock Pifuhd: Multi-level pixel-aligned implicit function for
  high-resolution 3d human digitization.
\newblock In {\em 2020 {IEEE/CVF} Conference on Computer Vision and Pattern
  Recognition, {CVPR} 2020, Seattle, WA, USA, June 13-19, 2020}, pages 81--90.
  Computer Vision Foundation / {IEEE}, 2020.

\bibitem{exemPose}
Katsuhito Sasaki, Keisuke Shiro, and Jun Rekimoto.
\newblock Exemposer: Predicting poses of experts as examples for beginners in
  climbing using a neural network.
\newblock In {\em Proceedings of the Augmented Humans International Conference,
  AHs 2020, Kaiserslautern, Germany, 16-17 March, 2020}, pages 18:1--18:9.
  {ACM}, 2020.

\bibitem{segal2009generalized}
Aleksandr Segal, Dirk Haehnel, and Sebastian Thrun.
\newblock Generalized-icp.
\newblock In {\em Robotics: science and systems}, volume~2, page 435. Seattle,
  WA, 2009.

\bibitem{Sigal2009HumanEvaSV}
Leonid Sigal, Alexandru~O. Balan, and Michael~J. Black.
\newblock Humaneva: Synchronized video and motion capture dataset and baseline
  algorithm for evaluation of articulated human motion.
\newblock {\em International Journal of Computer Vision}, 87:4--27, 2009.

\bibitem{Su2020RobustFusionHV}
Zhuo Su, Lan Xu, Zerong Zheng, Tao Yu, Yebin Liu, and Lu Fang.
\newblock Robustfusion: Human volumetric capture with data-driven visual cues
  using a rgbd camera.
\newblock In {\em ECCV}, 2020.

\bibitem{survey2022}
Yating Tian, Hongwen Zhang, Yebin Liu, and Limin Wang.
\newblock Recovering 3d human mesh from monocular images: {A} survey.
\newblock {\em CoRR}, abs/2203.01923, 2022.

\bibitem{ClimbingMR}
Marcel Tiator, Christian Geiger, Bastian Dewitz, Ben Fischer, Laurin Gerhardt,
  David Nowottnik, and Hendrik Preu.
\newblock Venga!: climbing in mixed reality.
\newblock In Stephan~G. Lukosch and Kai Kunze, editors, {\em Proceedings of the
  First Superhuman Sports Design Challenge: First International Symposium on
  Amplifying Capabilities and Competing in Mixed Realities, July 2-5, 2018,
  Delft, The Netherlands}, pages 9:1--9:8. {ACM}, 2018.

\bibitem{posendf}
Garvita Tiwari, Dimitrije Antic, Jan~Eric Lenssen, Nikolaos Sarafianos, Tony
  Tung, and Gerard Pons-Moll.
\newblock Pose-ndf: Modeling human pose manifolds with neural distance fields.
\newblock In {\em European Conference on Computer Vision ({ECCV})}, October
  2022.

\bibitem{Trumble2017TotalC3}
Matthew Trumble, Andrew Gilbert, Charles Malleson, Adrian Hilton, and John~P.
  Collomosse.
\newblock Total capture: 3d human pose estimation fusing video and inertial
  sensors.
\newblock In {\em BMVC}, 2017.

\bibitem{Vlasic2007PracticalMC}
Daniel Vlasic, Rolf Adelsberger, Giovanni Vannucci, John~C. Barnwell, Markus~H.
  Gross, Wojciech Matusik, and Jovan Popovi{\'c}.
\newblock Practical motion capture in everyday surroundings.
\newblock {\em ACM SIGGRAPH 2007 papers}, 2007.

\bibitem{3DPW}
Timo von Marcard, Roberto Henschel, Michael~J. Black, Bodo Rosenhahn, and
  Gerard Pons-Moll.
\newblock Recovering accurate 3d human pose in the wild using imus and a moving
  camera.
\newblock In {\em ECCV}, 2018.

\bibitem{SIP}
Timo {von Marcard}, Bodo Rosenhahn, Michael Black, and Gerard Pons-Moll.
\newblock Sparse inertial poser: Automatic 3d human pose estimation from sparse
  imus.
\newblock {\em Computer Graphics Forum 36(2), Proceedings of the 38th Annual
  Conference of the European Association for Computer Graphics (Eurographics)},
  pages 349--360, 2017.

\bibitem{MAED}
Ziniu Wan, Zhengjia Li, Maoqing Tian, Jianbo Liu, Shuai Yi, and Hongsheng Li.
\newblock Encoder-decoder with multi-level attention for 3d human shape and
  pose estimation.
\newblock In {\em 2021 {IEEE/CVF} International Conference on Computer Vision,
  {ICCV} 2021, Montreal, QC, Canada, October 10-17, 2021}, pages 13013--13022.
  {IEEE}, 2021.

\bibitem{replicating}
Emily Whiting, Nada Ouf, Liane Makatura, Christos Mousas, Zhenyu Shu, and
  Ladislav Kavan.
\newblock Environment-scale fabrication: Replicating outdoor climbing
  experiences.
\newblock In {\em Proceedings of the 2017 CHI Conference on Human Factors in
  Computing Systems}, pages 1794--1804. ACM, 2017.

\bibitem{woltring1974new}
HJ Woltring.
\newblock New possibilities for human motion studies by real-time light spot
  position measurement.
\newblock {\em Biotelemetry}, 1(3):132, 1974.

\bibitem{futurepose}
Erwin Wu and Hideki Koike.
\newblock Futurepose - mixed reality martial arts training using real-time 3d
  human pose forecasting with a {RGB} camera.
\newblock In {\em {IEEE} Winter Conference on Applications of Computer Vision,
  {WACV} 2019, Waikoloa Village, HI, USA, January 7-11, 2019}, pages
  1384--1392. {IEEE}, 2019.

\bibitem{ICON}
Yuliang Xiu, Jinlong Yang, Dimitrios Tzionas, and Michael~J. Black.
\newblock {ICON:} implicit clothed humans obtained from normals.
\newblock In {\em {IEEE/CVF} Conference on Computer Vision and Pattern
  Recognition, {CVPR} 2022, New Orleans, LA, USA, June 18-24, 2022}, pages
  13286--13296. {IEEE}, 2022.

\bibitem{Xu2020EventCapM3}
Lan Xu, Weipeng Xu, Vladislav Golyanik, Marc Habermann, Lu~Ming Fang, and
  Christian Theobalt.
\newblock Eventcap: Monocular 3d capture of high-speed human motions using an
  event camera.
\newblock {\em 2020 IEEE/CVF Conference on Computer Vision and Pattern
  Recognition (CVPR)}, pages 4967--4977, 2020.

\bibitem{MonoPerfCap}
Weipeng Xu, Avishek Chatterjee, Michael Zollh\"{o}fer, Helge Rhodin, Dushyant
  Mehta, Hans-Peter Seidel, and Christian Theobalt.
\newblock Monoperfcap: Human performance capture from monocular video.
\newblock {\em ACM Transactions on Graphics (TOG)}, 37(2):27:1--27:15, 2018.

\bibitem{yang2021s3}
Ze Yang, Shenlong Wang, Siva Manivasagam, Zeng Huang, Wei-Chiu Ma, Xinchen Yan,
  Ersin Yumer, and Raquel Urtasun.
\newblock S3: Neural shape, skeleton, and skinning fields for 3d human
  modeling.
\newblock In {\em CVPR}, 2021.

\bibitem{PIP}
Xinyu Yi, Yuxiao Zhou, Marc Habermann, Soshi Shimada, Vladislav Golyanik,
  Christian Theobalt, and Feng Xu.
\newblock Physical inertial poser {(PIP):} physics-aware real-time human motion
  tracking from sparse inertial sensors.
\newblock In {\em {IEEE/CVF} Conference on Computer Vision and Pattern
  Recognition, {CVPR} 2022, New Orleans, LA, USA, June 18-24, 2022}, pages
  13157--13168. {IEEE}, 2022.

\bibitem{Yi2021TransPoseR3}
Xinyu Yi, Yuxiao Zhou, and Feng Xu.
\newblock Transpose: Real-time 3d human translation and pose estimation with
  six inertial sensors.
\newblock {\em ACM Trans. Graph.}, 40:86:1--86:13, 2021.

\bibitem{GLAMR}
Ye Yuan, Umar Iqbal, Pavlo Molchanov, Kris Kitani, and Jan Kautz.
\newblock {GLAMR:} global occlusion-aware human mesh recovery with dynamic
  cameras.
\newblock In {\em {IEEE/CVF} Conference on Computer Vision and Pattern
  Recognition, {CVPR} 2022, New Orleans, LA, USA, June 18-24, 2022}, pages
  11028--11039. {IEEE}, 2022.

\bibitem{EgoBody}
Siwei Zhang, Qianli Ma, Yan Zhang, Zhiyin Qian, Taein Kwon, Marc Pollefeys,
  Federica Bogo, and Siyu Tang.
\newblock Egobody: Human body shape and motion of interacting people
  from head-mounted devices.
\newblock In Shai Avidan, Gabriel Brostow, Moustapha Ciss{\'e}, Giovanni~Maria
  Farinella, and Tal Hassner, editors, {\em Computer Vision -- ECCV 2022},
  pages 180--200, Cham, 2022. Springer Nature Switzerland.

\bibitem{LEMO}
Siwei Zhang, Yan Zhang, Federica Bogo, Marc Pollefeys, and Siyu Tang.
\newblock Learning motion priors for 4d human body capture in 3d scenes.
\newblock In {\em 2021 {IEEE/CVF} International Conference on Computer Vision,
  {ICCV} 2021, Montreal, QC, Canada, October 10-17, 2021}, pages 11323--11333.
  {IEEE}, 2021.

\bibitem{Zhang2013FromAT}
Weiyu Zhang, Menglong Zhu, and Konstantinos~G. Derpanis.
\newblock From actemes to action: A strongly-supervised representation for
  detailed action understanding.
\newblock {\em 2013 IEEE International Conference on Computer Vision}, pages
  2248--2255, 2013.

\bibitem{Zhou16a}
Xiaowei Zhou, Menglong Zhu, Spyridon Leonardos, Konstantinos~G Derpanis, and
  Kostas Daniilidis.
\newblock {Sparseness Meets Deepness: 3D Human Pose Estimation from Monocular
  Video}.
\newblock In {\em Computer Vision and Pattern Recognition (CVPR)}, 2016.

\end{thebibliography}
}

\end{document}